\documentclass[letterpaper, 10 pt, conference]{ieeeconf}  % 

\IEEEoverridecommandlockouts                   
\usepackage{graphics}
\usepackage{epsfig}
\usepackage{mathptmx}
\usepackage{times}
\usepackage{amsmath, amssymb, graphicx, algorithm, amsfonts, algpseudocode, etoolbox, tikz, mathdots, yhmath, cancel}

\let\labelindent\relax   
\usepackage{enumitem}

\title{\LARGE \bf
Adaptive Trajectory Optimization for Task-Specific Human-Robot Collaboration
}

\author{Hamed Rahimi Nohooji$^{1,*}$ and Holger Voos$^{1}$% <-this % stops a space
\thanks{*This research was funded in whole, or in part, by the
Luxembourg National Research Fund (FNR), COSAMOS
Project, ref. IC22/IS/17432865/COSAMOS. 
}% <-this % stops a space
\thanks{$^{1}$Hamed Rahimi Nohooji, and Holger Voos are with the Automation and Robotics Research Group, Interdisciplinary Centre for Security, Reliability and Trust, University of Luxembourg,
1855 Luxembourg, Luxembourg
        {\tt\small hamed.rahimi@uni.lu; holger.voos@uni.lu}}%
 }

\newtheorem{lemma}{Lemma}

\begin{document}

\maketitle
\thispagestyle{empty}
\pagestyle{empty}

\begin{abstract}
This paper proposes a task-specific trajectory optimization framework for human-robot collaboration, enabling adaptive motion planning based on human interaction dynamics. Unlike conventional approaches that rely on predefined desired trajectories, the proposed framework optimizes the collaborative motion dynamically using the inverse differential Riccati equation, ensuring adaptability to task variations and human input. The generated trajectory serves as the reference for a neuro-adaptive PID controller, which leverages a neural network to adjust control gains in real time, addressing system uncertainties while maintaining low computational complexity. The combination of trajectory planning and the adaptive control law ensures stability and accurate joint-space tracking without requiring extensive parameter tuning. Numerical simulations validate the proposed approach.
\end{abstract}

 % \begin{IEEEkeywords}
 % Inverse Differential Riccati Equation (iDRE), Human-Robot Collaboration, Neuro-Adaptive PID Control, Trajectory Optimization
 % \end{IEEEkeywords}

\section{Introduction}
Human-robot collaboration (HRC) is becoming increasingly important across fields such as manufacturing, healthcare, and service industries, where robots must operate in dynamic environments while interacting with humans. Unlike traditional robotic systems that follow predefined trajectories, collaborative robots require adaptability to human actions and environmental variations in real time. This necessitates control strategies that balance trajectory optimization and adaptive control, ensuring both precision and flexibility in interaction \cite{li2013human}. Conventional trajectory-based control methods often rely on fixed reference trajectories, limiting their ability to adjust to human guidance \cite{spyrakos2020minimally}. As a result, research has increasingly focused on integrating optimization-based trajectory planning with adaptive control mechanisms to enhance real-time robot performance.

Recent advancements have explored different strategies to improve adaptability and robustness in HRC. Optimization-based trajectory planning methods aim to generate trajectories that anticipate human actions while optimizing safety and efficiency. Decré et al. \cite{decre2014optimization} proposed a parametric modeling approach for trajectory generation, leveraging constrained optimization to minimize human-robot interaction forces. Jain et al. \cite{jain2020anticipatory} introduced CoMOTO, a framework for motion optimization that accounts for human motion prediction using a multi-objective cost function. Luo et al. \cite{luo2020adaptive} extended this concept by integrating trajectory adaptation with adaptive impedance control to improve interaction performance. Additionally, Wu et al. \cite{wu1706learning} developed a learning-based approach using Markov Decision Processes to maximize task success probability while minimizing associated costs. Zakaria et al. \cite{hani2021general} proposed a flexible decision-making framework that separates task execution from collaboration performance, enabling adaptive performance metrics. Rozo et al. \cite{rozo2015learning} explored learning from demonstrations to teach cooperative behaviors, incorporating probabilistic encoding and variable impedance control. These works address critical aspects of HRC, including trajectory optimization, task allocation, decision-making flexibility, and learning-based adaptation. However, while these approaches improve specific aspects of collaboration, they do not provide a unified solution that simultaneously optimizes trajectory generation and control adaptation.

Adaptive control techniques have also been widely studied for HRC, focusing on enhancing system robustness against uncertainties. Impedance control, originally introduced to regulate interaction forces, remains a common approach \cite{hogan1985impedance}. While effective, most implementations use fixed impedance parameters that assume constant interaction dynamics, limiting adaptability to real-time variations \cite{yousefizadeh2019trajectory}. To address this limitation, time-varying impedance models have been introduced, enabling robots to dynamically adjust their compliance based on task demands \cite{liu2020optimized}. Learning-based control strategies, such as reinforcement learning and neural-adaptive controllers, further enhance adaptability but often come at the cost of increased computational complexity, making them less practical for real-time applications \cite{ge2014impedance, li2016framework}.

Despite the advancements in trajectory optimization and adaptive control, existing methods have notable limitations. Many trajectory planning approaches rely on predefined reference paths, reducing real-time adaptability in dynamic human-robot interactions \cite{modares2015optimized}. In contrast, adaptive control methods improve compliance and robustness but do not inherently optimize trajectories, leading to suboptimal motion efficiency. Furthermore, approaches based on high-dimensional optimization or deep learning often introduce computational challenges that hinder real-time implementation. Addressing these limitations requires a framework that integrates trajectory optimization and adaptive control into a unified strategy that ensures both computational efficiency and stability.

This paper presents a task-specific human-robot collaboration framework that combines trajectory optimization with adaptive control, providing a computationally efficient solution for real-time applications. The framework integrates time-varying impedance-based interaction modeling to regulate human-robot interaction forces dynamically. It employs the inverse differential Riccati equation (iDRE) to optimize trajectories, eliminating the need for predefined reference paths while ensuring real-time adaptability. To achieve precise trajectory tracking, a neuro-adaptive PID controller with RBF neural network-based gain adaptation is introduced, ensuring low computational cost while maintaining stability. Unlike conventional PMP-based trajectory optimization, which requires solving a two-point boundary value problem \cite{korayem2008trajectory}, iDRE offers a closed-form, real-time optimization method that significantly reduces computational overhead. The neuro-adaptive PID control strategy eliminates the need for manual tuning, leveraging RBFNN-based adaptation to enhance flexibility.

The key contributions of this work are as follows:
\begin{itemize}
    \item The development of a unified HRC framework that integrates trajectory optimization and adaptive control, ensuring real-time adaptability without requiring predefined reference trajectories.
    \item The incorporation of time-varying impedance parameters to improve interaction compliance, enabling the robot to dynamically adjust its response based on task demands.
  \item The establishment of a direct connection between the optimized trajectory, derived from the human-robot collaboration model, and a computationally efficient, adaptive control strategy, facilitating smooth integration into the control framework while preserving stability.

\end{itemize}

The remainder of the paper is organized as follows: Section II presents the system modeling and problem definition. Section III derives the optimal task-specific trajectory by integrating human impedance dynamics with robot compliance, ensuring adaptive motion generation. Section IV describes the neuro-adaptive PID control strategy for accurate trajectory tracking. Section V validates the proposed framework through numerical simulations. Finally, Section VI provides a discussion of the findings and concludes the paper.

 \section{System Modeling and Problem Definition}

\subsection{System Description}
\label{subsec_description}
The robot's kinematics in task space are described by:
\begin{equation}
x = \Phi(q),
\label{eq-phi}
\end{equation}
where \( q(t) \in \mathbb{R}^n \) is the vector of generalized joint coordinates, and \( x(t) \in \mathbb{R}^{n_e} \) represents the Cartesian position of the end-effector, with \( n \) being the number of joints and \( n_e \) the dimension of the task space.

The robot dynamics in the joint space are expressed as:
\begin{equation}  \label{robot_joint}
\mathbf{M_r}(q)\ddot{q}(t) + \mathbf{C_r}(q,\dot{q})\dot{q}(t) + \mathbf{G_r}(q) = \tau(t) + \mathbf{J}^T(q) f_h(t),
\end{equation}
where \( \mathbf{M_r}(q) \in \mathbb{R}^{n \times n} \) is the positive-definite mass matrix, \( \mathbf{C_r}(q,\dot{q}) \in \mathbb{R}^{n \times n} \) represents the Coriolis and centrifugal forces, \( \mathbf{G_r}(q) \in \mathbb{R}^n \) is the gravitational force, and \( \mathbf{J}(q) \in \mathbb{R}^{n_e \times n} \) is the Jacobian matrix. The input torque is \( \tau(t) \in \mathbb{R}^n \), and \( f_h(t) \) is the human-applied force.

% By transforming the joint-space dynamics \eqref{robot_joint} into Cartesian space using \(\dot{x}(t) = \mathbf{J}(q)\dot{q}(t)\) and \(\ddot{x}(t) = \dot{\mathbf{J}}(q)\dot{q}(t) + \mathbf{J}(q)\ddot{q}(t)\), the Cartesian-space robot dynamics become:
% \begin{equation}   \label{robot_cartesian}
% \mathbf{M_c}(q)\ddot{x}(t) + \mathbf{C_c}(q,\dot{q})\dot{x}(t) + \mathbf{G_c}(q) = u(t) + u_h(t),
% \end{equation}
% where \( \mathbf{M_c}(q) = \mathbf{J}^{-T}(q) \mathbf{M_r}(q) \mathbf{J}^{-1}(q) \), \( \mathbf{C_c}(q,\dot{q}) = \mathbf{J}^{-T}(q) \left(\mathbf{C_r}(q,\dot{q}) - \mathbf{M_r}(q) \mathbf{J}^{-1}(q) \dot{\mathbf{J}}(q)\right) \mathbf{J}^{-1}(q) \), \( \mathbf{G_c}(q) = \mathbf{J}^{-T}(q) \mathbf{G_r}(q) \), \( u(t) = \mathbf{J}^{-T}(q) \tau(t) \), and \( u_h(t) =  f_h(t) \).

To guide the interaction, we employ an impedance model that defines the desired behavior of the system:
\begin{equation} \label{impedance_model} 
\mathbf{M_\text{imp}}(t)\ddot{x}_\text{imp}(t) + \mathbf{B_\text{imp}}(t) \dot{x}_\text{imp}(t) + \mathbf{K_\text{imp}}(t) x_\text{imp}(t) =  f_h(t),   
\end{equation}

\noindent where \( \mathbf{M_\text{imp}}(t) \), \( \mathbf{B_\text{imp}}(t) \), and \( \mathbf{K_\text{imp}}(t) \) are the time-varying impedance parameters: mass, damping, and stiffness, respectively. \( x_\text{imp}(t) \) represents the output of the prescribed impedance model. The impedance model is designed to regulate the interaction forces, ensuring smooth collaboration between the human and the robot by modulating the system’s response to applied forces.

The human impedance model describes the human-applied force \( f_h(t) \) in response to the robot's motion. The model can be expressed as:
\begin{equation} \label{human_imp} 
\left(\mathbf{K_d}(t) s + \mathbf{K_p}(t)\right) f_h(t) = \mathbf{k_e}(t) x_d(t),
\end{equation}
where \( s \) is the Laplace operator. \( \mathbf{K_d}(t) \) and \( \mathbf{K_p}(t) \) represent time-varying human damping and stiffness gains, respectively. \( \mathbf{k_e}(t) \) is the human control gain, and \( x_d(t) \) is the desired human trajectory. This model reflects the adaptive nature of human control during interaction.

\noindent
\textbf{Property 1} \cite{Lee1998}, \cite{Slotine1991}. The mass matrix \( \mathbf{M_r}(q) \) is symmetric, positive definite, and its norm is bounded by constants \( \alpha_m > 0 \) and \( \alpha_M > 0 \), such that
\(
\alpha_m \leq \| \mathbf{M_r}(q) \| \leq \alpha_M.
\)
Additionally, the Jacobian matrix \( \mathbf{J}(q) \) is bounded, i.e., there exists a positive constant \( J_{\max} \) such that
\(
\|\mathbf{J}(q)\| \leq J_{\max}.
\)

\noindent
\textbf{Property 2} \cite{Slotine1991}, \cite{Lewis2003}. The matrix \( \dot{\mathbf{M_r}}(q) - 2\mathbf{C_r}(q,\dot{q}) \) is skew-symmetric, i.e., for any vector \( \lambda \in \mathbb{R}^n \), \( \lambda^T \left( \dot{\mathbf{M_r}}(q) - 2\mathbf{C_r}(q,\dot{q}) \right) \lambda = 0 \). Furthermore, there exist constants \( \eta > 0 \) and \( \delta > 0 \), such that \( \|\mathbf{C_r}(q,\dot{q})\| \leq \eta \|\dot{q}(t)\| \) and \( \|\mathbf{G_r}(q)\| \leq \delta \).

\subsection{Problem Statement}

This work focuses on physical human-robot collaboration, where the robot’s motion is not predefined but adaptively influenced by human-applied forces. The objective is to develop a structured framework that optimizes motion planning and ensures stable trajectory tracking while maintaining computational efficiency. \textit{
The main challenge lies in integrating trajectory generation and adaptive control into a unified approach.} The framework must address the following key aspects:
\begin{itemize}
    \item The robot’s motion should be dynamically optimized based on human interaction forces rather than following predefined references. This requires generating an adaptive trajectory that balances compliance and task efficiency.
    \item  The planned trajectory must be accurately followed despite uncertainties in human input and system dynamics. A control strategy is needed that ensures stability while remaining computationally feasible for real-time applications.
\end{itemize}

\section{Task-Specific Assistive Human-Robot Collaboration} \label{SecTaskSpec}

This section formulates a unified interaction model that integrates the human impedance model with the robot’s compliant dynamics. By incorporating time-varying impedance parameters, the framework captures the adaptive nature of human input and optimizes collaboration without relying on predefined trajectories. The resulting system dynamics provide a structured foundation for trajectory generation, ensuring real-time adaptability and smooth interaction.

To model the human-robot interaction, we start by describing the human-generated force, which is time-varying and adaptive to the robot’s motion. The human impedance model (\ref{human_imp}) can be rewritten in the time domain as:
\begin{equation} \label{4} 
\mathbf{K_d}(t) \dot{f}_h(t) + \mathbf{K_p}(t) f_h(t) = \mathbf{k_e}(t) x_d(t).
\end{equation}
This can be equivalently rewritten as:

\begin{equation} \label{Human_timeD} 
\dot{f}_h(t) = \mathbf{A_h}(t) X_d(t) + \mathbf{B_h}(t) f_h(t),
\end{equation}

\noindent
where \( \mathbf{A_h}(t) = \mathbf{k_e(t)} \mathbf{K}_{d1}(t) \), 
and \( \mathbf{B_h}(t) = -\mathbf{K_d}(t)^{-1} \mathbf{K_p}(t) \), 
with \( \mathbf{K}_{d1}(t) = \begin{bmatrix} \mathbf{K_d}(t)^{-1} & 0 \end{bmatrix} \),
and 
\( X_d(t) = \begin{bmatrix} x_d(t)^T & \dot{x}_d(t)^T \end{bmatrix}^T \).

Now, define the impedance state vector \(\xi(t) = \begin{bmatrix} x_\text{imp}(t)^T & \dot{x}_\text{imp}(t)^T \end{bmatrix}^T\), where \(x_\text{imp}(t)\) and \(\dot{x}_\text{imp}(t)\) represent the desired position and velocity of the impedance behavior. The impedance model (\ref{impedance_model}) can then be expressed as:
\begin{equation} \label{ImpState} 
\dot{\xi}(t) = \mathbf{A_\xi}(t) \xi(t) + \mathbf{B_\xi}(t) u(t),
\end{equation}
where \(u(t) = \mathbf{M_\text{imp}}(t)^{-1}  f_h(t)\), and the system matrices \(\mathbf{A_\xi}(t)\) and \(\mathbf{B_\xi}(t)\) are defined as:
$
\mathbf{A_\xi}(t) = \begin{bmatrix} 0 & \mathbf{I}_{n \times n} \\ -\mathbf{M_\text{imp}}(t)^{-1} \mathbf{K_\text{imp}}(t) & -\mathbf{M_\text{imp}}(t)^{-1} \mathbf{B_\text{imp}}(t) \end{bmatrix}$, and 
$\quad
\mathbf{B_\xi}(t) = \begin{bmatrix} 0 \\ \mathbf{I}_{n \times n} \end{bmatrix}.
$

% \noindent
It is important to note that the human's desired trajectory \(x_d(t)\) coincides with the reference trajectory of the impedance model, i.e., \(\xi = X_d\). This assumption physically reflects the natural alignment of the human's control efforts with the robot's compliant behavior. By linking the robot’s impedance reference with the human’s desired motion, the model effectively captures the interactive nature of human-robot collaboration, where the human adjusts their input to guide the robot along a shared trajectory.

\noindent
By combining the human and impedance models, i.e., \eqref{Human_timeD}, and \eqref{ImpState}, we obtain:
\begin{equation} \label{xdot} 
\dot{X}(t) = \mathbf{A}(t) X(t) + \mathbf{B}(t) u(t),
\end{equation}
where \(X(t) = \begin{bmatrix} \xi(t)^T & f_h(t)^T \end{bmatrix}^T\), \(\mathbf{A}(t) = \begin{bmatrix} \mathbf{A_\xi}(t) & 0 \\ \mathbf{A_h}(t) & \mathbf{B_h}(t) \end{bmatrix}\), and \(\mathbf{B}(t) = \begin{bmatrix} \mathbf{B_\xi}(t) \\ 0 \end{bmatrix}\). This unified dynamic model represents the interaction between the human and the robot, facilitating task-specific collaboration.

The objective is to minimize the following cost function:
\begin{equation} \label{ZEqnNum481895} 
\begin{split}
E = \frac{1}{2} \int_{t_0}^{t_f} & \left( X^T(t) \mathbf{Q} X(t)  + u^T(t) \mathbf{R} u(t) \right) dt,
\end{split}
\end{equation}
where \(\mathbf{Q} \in \mathbb{R}^{n \times n}\) and \(\mathbf{R} \in \mathbb{R}^{m \times m}\) are constant weight matrices, with \(\mathbf{Q} \geq 0\) and \(\mathbf{R} > 0\). Then, following \cite{rahimi2020optimal}, and considering the system dynamics \eqref{xdot} the optimal HRC trajectory is obtained as:
\begin{equation} \label{X_optimal}
\begin{aligned}
\dot{X}^*(t) &= \left( \mathbf{A}(t) - \mathbf{B}(t) \mathbf{R}^{-1} 
\mathbf{B}^T(t) \mathbf{P}^{-1}(t) \right) X^*(t)  \\
&\quad + \mathbf{B}(t) \mathbf{R}^{-1} \mathbf{B}^T(t) \mathbf{P}^{-1}(t) V(t),
\end{aligned}
\end{equation}
where \(\mathbf{P}(t)\) and \(V(t)\) are solutions to the inverse matrix differential Riccati equation:
\begin{equation} \label{P} 
\begin{aligned}
\dot{\mathbf{P}}(t) &= \mathbf{A}(t) \mathbf{P}(t) + \mathbf{P}(t) \mathbf{A}^T(t) 
+ \mathbf{P}(t) \mathbf{Q} \mathbf{P}(t)  \\
&\quad - \mathbf{P}(t) \mathbf{B}(t) \mathbf{R}^{-1} \mathbf{B}^T(t) \mathbf{P}(t),
\end{aligned}
\end{equation}
and the vector differential equation for \(V(t)\):
\begin{equation} \label{V} 
\dot{V}(t) = (\mathbf{A}(t) + \mathbf{P}(t) \mathbf{Q}) V(t).
\end{equation}
Finally, the set of above equations can be solved using either the initial or final boundary conditions: \( t = t_0: \mathbf{P}(t_0) = 0, V(t_0) = X(t_0) \) or \( t = t_f: \mathbf{P}(t_f) = 0, V(t_f) = X(t_f) \).

Note that the generated trajectory 
$\dot{X}^*(t)$ serves as the optimal reference for the subsequent control design, ensuring that the robot follows a task-specific motion plan derived from the human-robot interaction model.

\section{Robot-Specific Control Design}
With the desired trajectory for human-robot collaboration established in the previous section, it is essential to ensure that the robot precisely tracks this trajectory throughout the interaction. Effective trajectory tracking is important for maintaining the desired behavior of the system and ensuring successful task execution. While there are several control methods available for trajectory tracking, some approaches are highly complex and computationally intensive. On the other hand, simpler methods may fail to provide stability guarantees, particularly in the presence of uncertainties.

To address these challenges, this work develops a neuro-adaptive PID control approach. This control method offers a balanced solution that ensures stability while remaining computationally efficient. The neuro-adaptive component dynamically adjusts the PID gains in response to system variations, making the approach both robust and adaptable to the varying conditions encountered in human-robot collaboration tasks.

Define the position error \( e(t) \in \mathbb{R}^n \), as \( e(t) = q_d(t) - q(t) \), where \( q_d(t) \) is the desired joint space trajectory, derived from the planned path in the previous section, and \( q(t) \) is the actual robot joint position. By ensuring that \( e(t) \to 0 \), we can confirm that the robot successfully tracks the generated trajectory.
To capture the tracking error in a form suitable for analysis, we define the generalized commutative error variable \( e_c(t) \) as:
\begin{equation} \label{eq:E1}
e_c(t) = 2\zeta e(t) + \zeta^2 \int_0^t e(\theta) \, d\theta + \frac{d}{dt}e(t),
\end{equation}
where \( \zeta > 0 \).
 This structure aligns with PID control, where the derivative enhances system dynamics, the integral removes steady-state error, and the proportional improves response tracking, addressing the tracking error and stabilizing the system as shown in the following lemma.

\begin{lemma} \cite{chen2021tracking}
Given the commutative variable \( e_c(t) \)  in \eqref{eq:E1}, if \( e_c(t) \to 0 \) as \( t \to \infty \), then the tracking errors \( e(t) \), \( \dot{e}(t) \), and their integrals are bounded and converge to zero as time progresses.
\end{lemma}

\noindent Next, to establish the stability of the control system, we present the following lemma based on the Lyapunov stability approach:

\begin{lemma} \label{Lemma_V} \cite{rahimi2018neural}
Consider the Lyapunov function \( V(t) = \frac{1}{2} \epsilon(t)^T \mathbf{Q} \epsilon(t) + \frac{1}{2} \tilde{\theta}(t)^T  \tilde{\theta}(t) \), where  \( \tilde{\theta}(t) = \theta^* - \hat{\theta}(t) \) with constants \( \theta^* \in \mathbb{R}^m \) and \( \hat{\theta}(t) \in \mathbb{R}^m \), and the matrices \( \mathbf{Q}(t) = \mathbf{Q}^T(t) > 0 \). If 
\begin{equation} 
\dot{V}(t) \leq -\alpha_1 V(t) + \alpha_2
\end{equation}
for positive constants \( \alpha_1 \) and \( \alpha_2 \), then the error \( \epsilon(t) \) and the uncertainty \( \tilde{\theta}(t) \) remain bounded.
\end{lemma}

The proposed PID-like control input  is formulated as:
\begin{equation} \label{eq_tau1}
\tau = \left(k_{rc} + k_R(t)\right)e_c(t),
\end{equation}
where \( k_{rc} \) is a constant positive control gain, and \( k_R(t) \) represents the time-varying gain. The \( e_c(t) \) is the commutative error defined earlier, capturing the PID concept.
The update law for \( k_R(t) \) is given by:
\begin{equation} \label{kappa}
k_R(t) = \alpha \hat{\theta}(t)^T \phi(z),
\end{equation}
where \( \alpha > 0 \) is a control constant, \( \hat{\theta}(t) \) is the adaptive weight vector, and \( \phi(z) \) is the basis function vector of the neural network with input \( z \), where \( z = [e, \dot{e}, e_c]^T \). 
The adaptive law for updating the weight vector \( \hat{\theta}(t) \) is defined as:
\begin{equation} \label{eq:hat{theta}}
\dot{\hat{\theta}}(t) = \alpha\|e_c(t)\|^2 \phi(z) - \sigma \hat{\theta}(t),
\end{equation}
where \( \sigma > 0 \) is a small positive constant.

\textbf{Theorem 1.}
\textit{
Given the robot dynamics \eqref{robot_joint} and Properties 1, and 2, and assuming the control law in \eqref{eq_tau1} with time-varying gain updating as in \eqref{kappa}, together with the adaptive law \eqref{eq:hat{theta}}, the following statements hold:
1) All signals in the closed-loop system are uniformly bounded.
2) The tracking error \( e(t) \) in the closed-loop system converges to a small neighborhood of zero as \( t \to \infty \), provided the design parameters are properly chosen.}

\begin{proof}
Consider the Lyapunov candidate function:
\begin{equation} \label{Lyap}
V(t) = \frac{1}{2} e_c(t)^T \mathbf{M}(q(t)) e_c(t) + \frac{1}{2} \tilde{\theta}(t)^T \tilde{\theta}(t),
\end{equation}
where \( \tilde{\theta}(t) = \theta^* - \hat{\theta}(t) \), and \( \theta^* \) is the ideal constant weight vector in the RBFNN approximation.

\noindent
Taking the time derivative of \( V(t) \), we get:
\begin{equation} \label{eq_vdot}
\begin{split}
\dot{V}(t) &= e_c(t)^T \mathbf{M}(q(t)) \dot{e_c}(t) + \frac{1}{2} e_c(t)^T \dot{\mathbf{M}}(q(t)) e_c(t) \\
&\quad + \tilde{\theta}(t)^T  \dot{\hat{\theta}}(t),
\end{split}
\end{equation}

\noindent
Using the robot dynamics \eqref{robot_joint} and the control law in \eqref{eq_tau1}, we obtain:
\begin{equation} \label{eq:ec_dyn}
\begin{aligned}
    \mathbf{M}(q) \dot{e}_c &= \mathbf{C}(q, \dot{q}) \dot{q} + \mathbf{G}(q) - \tau - \mathbf{J}^T(q) f_h \\
    &\quad + \mathbf{M}(q) \Bar{e}(t),
\end{aligned}
\end{equation}
where \( \Bar{e}(t) = \ddot{q}_d + 2\zeta \dot{e} + \zeta^2 e \).
Applying Young’s inequality, the following bounds hold:
\begin{equation} \label{eq:young1}
    e_c^T \mathbf{C}(q, \dot{q}) \dot{q} \leq \alpha \|e_c\|^2 \eta^2 \|\dot{q}\|^4 + \frac{1}{4\alpha},
\end{equation}
\begin{equation} \label{eq:young2}
    e_c^T (\mathbf{G}(q) - \mathbf{J}^T f_h) \leq \alpha \|e_c\|^2 (\delta + F)^2 + \frac{1}{4\alpha},
\end{equation}
where \( \|\mathbf{J}^T f_h\| \leq F \).
Additionally, for the dynamic terms:
\begin{equation} \label{eq:young3}
    e_c^T \mathbf{M}(q) \Bar{e} \leq \alpha \|e_c\|^2 \alpha_M^2 \|\Bar{e}\|^2 + \frac{1}{4\alpha},
\end{equation}
\begin{equation} \label{eq:young4}
    \frac{1}{2} e_c^T \dot{\mathbf{M}} e_c = e_c^T \mathbf{C}(q, \dot{q}) e_c 
    \leq \alpha \|e_c\|^2 \eta^2 \|\dot{q}\|^2 \|e_c\|^2 + \frac{1}{4\alpha}.
\end{equation}

\noindent
Substituting the bounds into the Lyapunov function derivative \eqref{eq_vdot}, we obtain:
\begin{equation} \label{eq_vdot_bound}
\begin{aligned}
    \dot{V}(t) &\leq -\alpha \|e_c(t)\|^2 \Lambda + \frac{1}{\alpha} + \tilde{\theta}^T  \dot{\hat{\theta}} \\
    &\quad - e_c^T (k_{rc} + k_R(t)) e_c.
\end{aligned}
\end{equation}

\noindent
Here, the term \( k_R(t) \) represents the adaptive gain updated by the neural network, and the function \( \Lambda(t) \) is defined as:
\begin{equation} \label{eq_lambda}
\begin{aligned}
    \Lambda(t) &= \eta^2 \|\dot{q}(t)\|^4 + (\delta + F)^2 + \alpha_M^2 \|\Bar{e}(t)\|^2 \\
    &\quad + \eta^2 \|\dot{q}(t)\|^2 \|e_c(t)\|^2. 
\end{aligned}
\end{equation}

\noindent
Substituting \( k_R(t) \) from \eqref{kappa} and the adaptive law from \eqref{eq:hat{theta}} into \eqref{eq_vdot_bound}, we obtain:
\begin{equation} \label{eq_vdot_final2}
\begin{aligned}
    \dot{V}(t) &\leq -\alpha \|e_c\|^2 \Lambda + \frac{1}{\alpha} - e_c^T k_{rc} e_c \\
    &\quad - \alpha e_c^T \hat{\theta}^T \phi(z) e_c + \tilde{\theta}^T  \left( \alpha \|e_c\|^2 \phi(z) - \sigma \hat{\theta} \right).
\end{aligned}
\end{equation}

To approximate the function \( \Lambda(t) \), we employ an RBFNN representation as $ \Lambda(t) = \theta^{*T} \phi(z) + \epsilon_\Lambda,$
where \( \theta^* \) is the ideal weight vector, \( \phi(z) \) is the basis function vector, and \( \epsilon_\Lambda \) is the bounded approximation error satisfying \( |\epsilon_\Lambda| \leq \epsilon_b \). \\
Using this approximation and noting that \( \tilde{\theta} = \theta^* - \hat{\theta} \), we apply the identity
$\tilde{\theta}^T \hat{\theta} = \tilde{\theta}^T \theta^* - \tilde{\theta}^T \tilde{\theta},$
with the bound
$    \tilde{\theta}^T \theta^* \leq \frac{1}{2} (\tilde{\theta}^T \tilde{\theta} + \theta^{*T} \theta^*).$
Then, applying algebraic manipulations, \eqref{eq_vdot_final2} can be rewritten as:
\begin{equation} \label{eq_vdot_final7}
\begin{aligned}
    \dot{V} &\leq - k_{rc} \|e_c\|^2 - \frac{\sigma}{2} \|\tilde{\theta}\|^2 + \alpha \|e_c\|^2 \epsilon_\Lambda \\
    &\quad + \alpha \|e_c\|^2 \epsilon_M + \frac{\sigma}{2} \|\theta^*\|^2 + \frac{1}{\alpha}.
\end{aligned}
\end{equation}

Accordingly, \eqref{eq_vdot_final7} can be rewritten as:
\begin{equation} \label{eq_vdot_final8}
    \dot{V} \leq - v_1 V + v_2,
\end{equation}
where $ v_1 = \min \left\{ 2 k_{rc} / \alpha_M, \sigma \right\},$ and $ v_2 = \alpha \|e_c\|^2 \epsilon_M + \frac{\sigma}{2} \|\theta^*\|^2 + \frac{1}{\alpha}.$

Using \eqref{Lyap}, Property 1, and Lemma 2, it follows that \( e_c(t) \) and \( \tilde{\theta}(t) \) are uniformly ultimately bounded. Applying Lemma 1, the tracking error \( e(t) \) is also bounded and thus the joint position \( q(t) \) remains bounded. Given the boundedness of \( \tilde{\theta}(t) \), the estimated weight \( \hat{\theta}(t) \) is also bounded. Moreover, since the basis function \( \phi(z) \) is bounded, the adaptive gain \( k_R(t) \) remains bounded. 
Thus, the control input \( \tau(t) \) is bounded, ensuring the boundedness of all closed-loop signals.
\end{proof}
Note that the developed neuro-adaptive PID control framework provides stability guarantees with a simple structure, minimal computational cost, and few tunable parameters. This makes it highly suitable for real-time human-robot collaboration tasks, even in uncertain environments.

Algorithm 1 outlines the proposed framework for task-specific human-robot collaboration, integrating trajectory optimization and adaptive control.

\begin{algorithm}[H]
\caption{Task-Specific Human-Robot Collaboration Framework}
\begin{algorithmic}
\State \textbf{Input:} System matrices \(\mathbf{A}(t)\), \(\mathbf{B}(t)\), weight matrices \(\mathbf{Q}\), \(\mathbf{R}\), control parameters \(\alpha, \sigma, \zeta, k_{rc}\), RBF parameters and initial weight vector \(\hat{\theta}(0)\)
\State \textbf{Initialize:} Desired initial and final states \( X_0, X_f \); Initialize \( \mathbf{P}(t), V(t) \) using the boundary conditions from iDRE

\State \textbf{Phase 1: Task-Specific Trajectory Optimization}
\While{$t < t_{final}$}
    \State Solve the inverse differential Riccati equation \eqref{P} and the vector differential equation  \eqref{V} to obtain \(\mathbf{P}(t)\) and \(V(t)\)
    \State Compute the optimal trajectory \( X^*(t) \) in \eqref{X_optimal} and update the system state \( X(t) \) using the state-space equation  \eqref{xdot}
\EndWhile

\State \textbf{Phase 2: Robot-Specific Tracking Control }
\While{$t < t_{final}$}
    \State Compute the tracking error \( e(t) = q_d(t) - q(t) \) and the commutative error \( e_c(t) \) in  \eqref{eq:E1}
    \State Update the adaptive weight vector \( \hat{\theta}(t) \) using the adaptive law  \eqref{eq:hat{theta}}, update the time-varying gain \( k_R(t) \) using \eqref{kappa}, and apply the control law \( \tau(t) = (k_{rc} + k_R(t))e_c(t) \)
    \State Update the robot’s system state using the robot dynamics and control law
\EndWhile

\State \textbf{Output:} Optimized human-robot trajectory and control performance
\end{algorithmic}
\end{algorithm}

\section{Simulation Study}
\label{sec_simulation}
In this section, simulations are conducted to validate the overall performance of the proposed framework, including both the trajectory optimization and task-specific collaboration strategies. A two-link robot manipulator operating in the vertical plane is used for the simulation. The desired trajectory is generated using the human-robot collaboration model defined in Section III, and the tracking control is designed based on the neuro-adaptive PID approach described in Section IV.

The physical parameters of the robot manipulator are as follows: masses of the links \( m_1 = 5 \, \text{kg} \) and \( m_2 = 5 \, \text{kg} \), lengths of the links \( L_1 = 1 \, \text{m} \) and \( L_2 = 1 \, \text{m} \), with joint inertia values \( I_1 = \frac{L_1}{12} \) and \( I_2 = \frac{L_2}{12} \). The impedance model matrices are chosen as \( \mathbf{M}_{imp} = \begin{bmatrix} 5 & 1 \\ 1 & -3 \end{bmatrix} \), \( \mathbf{B}_{imp} = \begin{bmatrix} 20 & 0 \\ 5 & 15 \end{bmatrix} \), and \( \mathbf{K}_{imp} = \begin{bmatrix} 1.0 & 0.5 \\ 0 & 0 \end{bmatrix} \). For the human dynamics, the parameters are set as \( \mathbf{K}_d = 10 \, \text{I} \), \( \mathbf{K}_p = 2 \, \text{I} \), and \( \mathbf{K}_e = \text{I} \), where \( \text{I} \) is the identity matrix. The initial joint conditions are calculated using inverse kinematics based on the desired Cartesian position.
The robot's initial joint positions and velocities are \( q_1(0) = 0.5 \, \text{rad} \), \( q_2(0) = 1 \, \text{rad} \), with joint velocities \( \dot{q}_1(0) = 0 \, \text{rad/s} \), \( \dot{q}_2(0) = 0 \, \text{rad/s} \). The corresponding initial end-effector Cartesian position is \( x_0 = -0.50 \, \text{m} \), \( y_0 = 1 \, \text{m} \), and the final desired position is \( x_f = 0.8 \, \text{m} \), \( y_f = -0.6 \, \text{m} \), with zero terminal velocities. 
The control parameters are \( \zeta = 0.1 \), \( k_{rc} = 50 \), \( \alpha = 10 \), and \( \sigma = 0.1 \). Twenty RBFNN nodes are used with zero-initialized weights. The matrices \( \mathbf{Q} = \mathbf{I} \), and \( \mathbf{R} = \mathbf{I} \) . The initial values for \( V(t) \), \( P(t) \), and \( f_h \) are zero.

\begin{figure}[!t]
\centerline{\includegraphics[width=\linewidth]{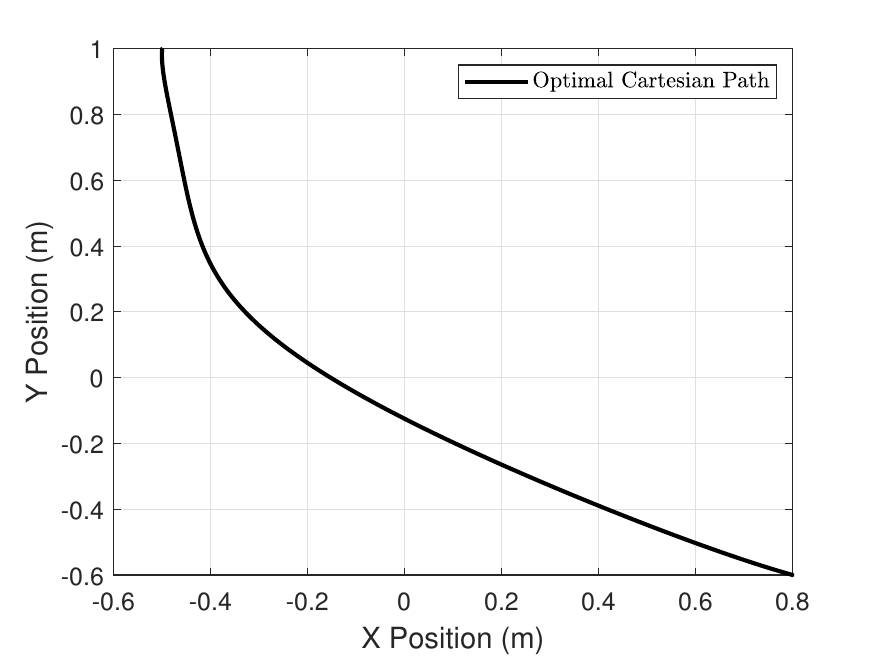}}
\caption{Cartesian trajectory in the $XY$ plane.}
\label{fig:1}
\end{figure}

\begin{figure}[!t]
\centerline{\includegraphics[width=\linewidth]{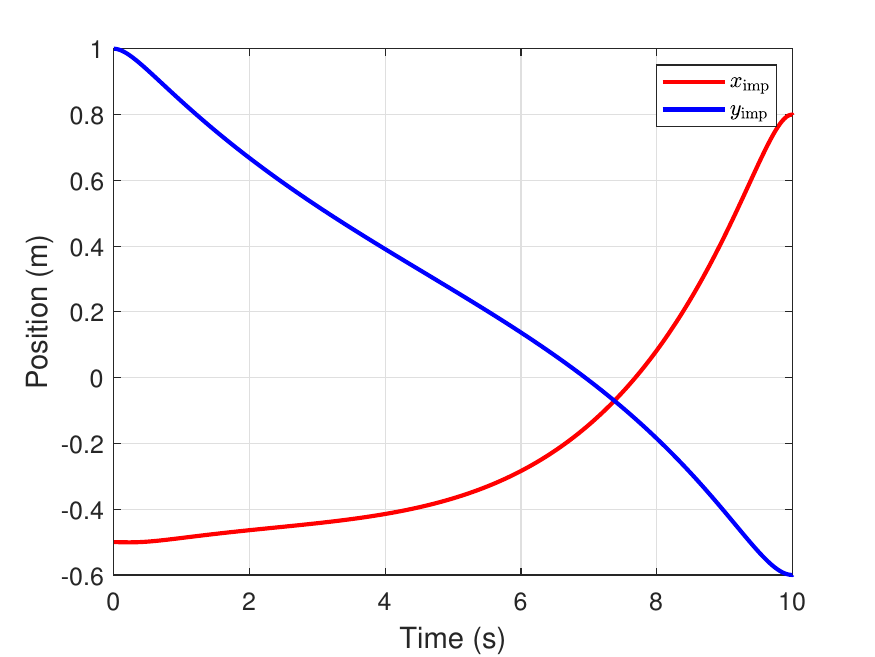}}
\caption{Positions in the Cartesian space: $x_\mathrm{imp}$ and $y_\mathrm{imp}$.}
\label{fig:2}
\end{figure}

\begin{figure}[!t]
\centerline{\includegraphics[width=\linewidth]{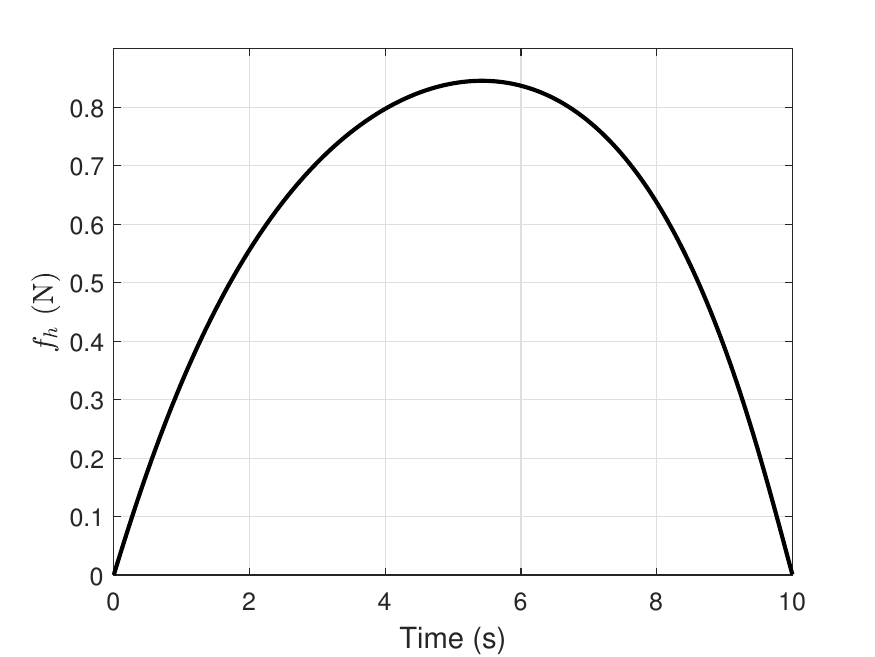}}
\caption{Interaction force $f_h$ during human-robot collaboration.}
\label{fig:3}
\end{figure}

\begin{figure}[!t]
\centerline{\includegraphics[width=\linewidth]{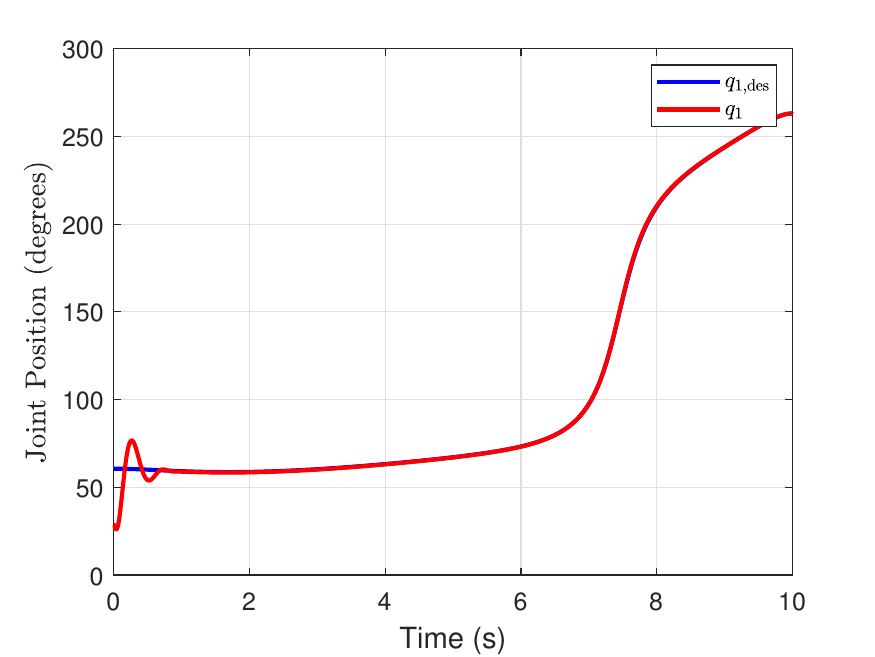}}
\caption{Desired and actual trajectories for the first joint.}
\label{fig:4}
\end{figure}

\begin{figure}[htbp]
\centerline{\includegraphics[width=\linewidth]{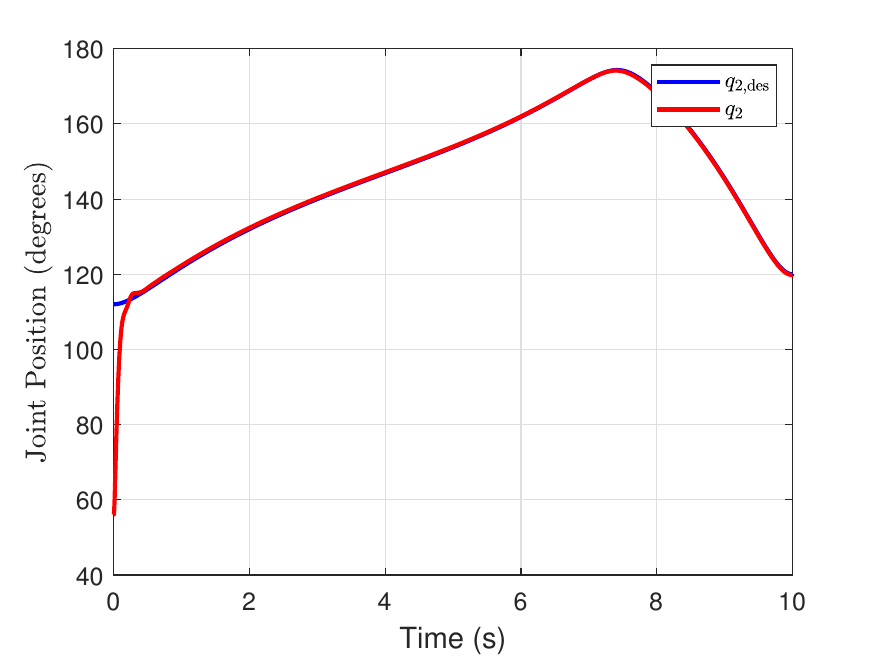}}
\caption{Desired and actual trajectories for the second joint.}
\label{fig:5}
\end{figure}

\begin{figure}[!t]
\centerline{\includegraphics[width=\linewidth]{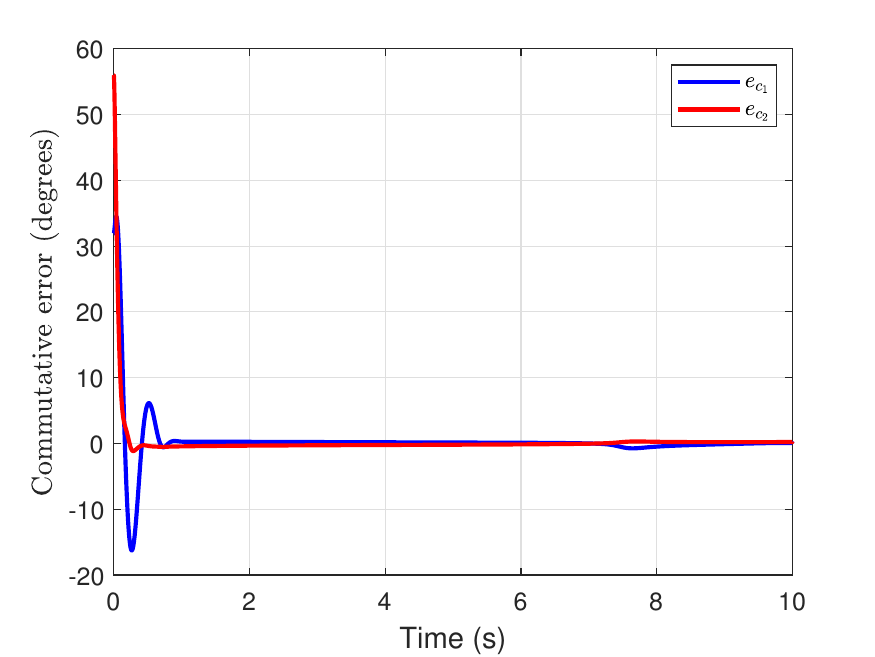}}
\caption{Trajectories of the commutative tracking errors  $e_{c_1}$ and $e_{c_2}$.}
\label{fig:6}
\end{figure}

Simulation results are presented in Figures \ref{fig:1} to \ref{fig:6}. 
Figures \ref{fig:1} to \ref{fig:3} focus on the task-specific human-robot collaboration.  The Cartesian trajectory of the robot's end-effector in the $x$ and $y$ directions is shown in Figure \ref{fig:1}, while the corresponding time-based trajectories are illustrated in Figure \ref{fig:2}. Figure \ref{fig:3} shows the interaction force $f_h$ during the collaboration. These results demonstrate the effectiveness of the proposed iDRE-based optimization for trajectory generation and collaboration.

The joint-space tracking performance is presented in Figures \ref{fig:4} to \ref{fig:6}. Figures \ref{fig:4} and \ref{fig:5} show the tracking performance of the first and second joint, respectively. Figure \ref{fig:6} illustrates the commutative tracking errors for both joints. The results indicate that the joint positions closely track the desired trajectories, with minimal tracking errors throughout the simulation.

The results confirm the effectiveness of the proposed framework in optimizing task-specific collaboration and ensuring joint-space tracking errors, demonstrating robust performance.

\section{Discussion and Conclusion}

The proposed framework integrates optimization-based trajectory planning with adaptive control to enhance HRC. By leveraging iDRE, the method optimizes time-varying collaboration dynamics, eliminating the need for a predefined desired trajectory. This enables the system to continuously adapt to human input while maintaining task efficiency. Additionally, the incorporation of a low-computation neuro-adaptive PID controller ensures accurate trajectory tracking with stability guarantees, making it suitable for real-time applications.

Compared to conventional methods, this framework achieves three main advantages. First, it unifies human impedance dynamics, robotic compliance behavior, and trajectory planning into a single, structured formulation. This direct integration enables a smooth transition between trajectory generation and control, reducing inconsistencies found in separately designed models. Second, the use of iDRE allows for trajectory optimization without solving a computationally expensive two-point boundary value problem, unlike PMP. This enhances computational efficiency while preserving real-time adaptability. Third, the neuro-adaptive PID control strategy ensures stability without requiring extensive parameter tuning, further improving practical implementation feasibility.

The primary novelty of this framework lies in the integration of iDRE, impedance modeling, and neuro-adaptive control into a coherent, task-specific optimization approach. The direct connection between human-robot interaction dynamics and the control law enables optimized motion generation while ensuring robust and efficient execution. Unlike prior works that apply these methods in isolation, this framework establishes a structured foundation where trajectory optimization and adaptive control complement each other naturally. This synergy not only enhances real-time adaptability but also simplifies implementation, making it a promising candidate for practical HRC applications.

In conclusion, the proposed framework provides a structured, optimization-based solution for task-specific human-robot collaboration, combining trajectory generation and control in a computationally efficient manner. The numerical results confirm its effectiveness in achieving stable and adaptive interaction. Future work will focus on experimental validation and extending the framework to more complex collaborative scenarios.

 \bibliographystyle{IEEEtran}

\end{document}